\documentclass[10pt,twocolumn,letterpaper]{article}
\usepackage{cvpr}              
\usepackage{graphicx}
\usepackage{amsmath}
\usepackage{amssymb}
\usepackage{booktabs}

\usepackage{caption}
\usepackage[nolist]{acronym}
\begin{acronym}[PHOENIX14\textbf{T} ]

\acrodefplural{rnn}[RNNs]{Recurrent Neural Networks}
\acrodefplural{cnn}[CNNs]{Convolutional Neural Networks}
\acrodefplural{hmm}[HMMs]{Hidden Markov Models}
\acrodefplural{gru}[GRUs]{Gated Recurrent Units}
\acrodefplural{crf}[CRFs]{Conditional Random Fields}
\acrodefplural{gan}[GANs]{Generative Adversarial Networks}
\acrodefplural{gpu}[GPUs]{Graphic Processing Units}

\acrodefplural{mdn}[MDNs]{Mixture Density Networks}

\acro{bsl}[BSL]{British Sign Language}
\acro{bleu}[BLEU]{Bilingual Evaluation Understudy}
\acro{blstm}[BLSTM]{Bidirectional Long Short-Term Memory}
\acro{bslcpt}[BSLCP\textbf{T}]{BSL Corpus \textbf{T}}
\acro{cnn}[CNN]{Convolutional Neural Network}
\acro{crf}[CRF]{Conditional Random Field}
\acro{cslr}[CSLR]{Continuous Sign Language Recognition}
\acro{ctc}[CTC]{Connectionist Temporal Classification}
\acro{c4a}[C4A]{Content4All}
\acro{dl}[DL]{Deep Learning}
\acro{dgs}[DGS]{German Sign Language - Deutsche Gebärdensprache}
\acro{dsgs}[DSGS]{Swiss German Sign Language - Deutschschweizer Geb\"ardensprache}
\acro{dtw}[DTW]{Dynamic Time Warping}
\acro{fc}[FC]{Fully Connected}
\acro{ff}[FF]{Feed Forward}
\acro{gan}[GAN]{Generative Adversarial Network}
\acro{gpu}[GPU]{Graphics Processing Unit}
\acro{gru}[GRU]{Gated Recurrent Unit}
\acro{gtpt}[G2PT]{Gloss to Pose Transformer}
\acro{hmm}[HMM]{Hidden Markov Model}
\acro{hpe}[HPE]{Hand Pose Enhancer}
\acro{isl}[ISL]{Irish Sign Language}
\acro{lstm}[LSTM]{Long Short-Term Memory}
\acro{mha}[MHA]{Multi-Headed Attention}
\acro{mtc}[MTC]{Monocular Total Capture}
\acro{mse}[MSE]{Mean Squared Error}
\acro{mdn}[MDN]{Mixture Density Network}
\acro{mdgs}[mDGS]{meineDGS}
\acro{mdgsv}[mDGS-V]{meineDGS-Variants}
\acro{mdgst}[mDGS\textbf{T}]{meineDGS\textbf{T}}
\acro{mdgsth}[mDGS\textbf{T}-\textbf{H}]{meineDGS\textbf{T}-\textbf{HARD}}
\acro{mdgste}[mDGS\textbf{T}-\textbf{E}]{meineDGS\textbf{T}-\textbf{EASY}}

\acro{nmt}[NMT]{Neural Machine Translation}
\acro{nlp}[NLP]{Natural Language Processing}
\acro{ph12}[PHOENIX12]{RWTH-PHOENIX-Weather-2012}
\acro{ph14}[PHOENIX14]{RWTH-PHOENIX-Weather-2014}
\acro{ph14t}[PHOENIX14\textbf{T}]{RWTH-PHOENIX-Weather-2014\textbf{T}}
\acro{pof}[POF]{Part Orientation Field}
\acro{paf}[PAF]{Part Affinity Field}
\acro{pttt}[P2TT]{Pose to Text Transformer}
\acro{relu}[RELU]{Rectified Linear Units}
\acro{rnn}[RNN]{Recurrent Neural Network}
\acro{rouge}[ROUGE]{Recall-Oriented Understudy for Gisting Evaluation}
\acro{sgd}[SGD]{Stochastic Gradient Descent}
\acro{sla}[SLA]{Sign Language Assessment}
\acro{slr}[SLR]{Sign Language Recognition}
\acro{slt}[SLT]{Sign Language Translation}
\acro{slp}[SLP]{Sign Language Production}
\acro{smt}[SMT]{Statistical Machine Translation}

\acro{ttgt}[T2GT]{Text to Gloss Transformer}
\acro{ttpt}[T2PT]{Text to Pose Transformer}
\acro{ttp}[T2P]{Text to Pose}
\acro{ttgtp}[T2G2P]{Text to Gloss to Pose}
\acro{tts}[TTS]{Text to Speech}
\acro{wer}[WER]{Word Error Rate}

\end{acronym}
\usepackage{dblfloatfix}

\usepackage[dvipsnames]{xcolor}

\newcommand{\FrameSelectionNet}{\textsc{FS-Net}}
\newcommand{\FrameSelectionNetLong}{Frame Selection Network}

\def\B{\fontseries{b}\selectfont}
\usepackage{multirow}
\DeclareMathOperator*{\argmax}{argmax}

\newcommand{\methodName}{\textsc{SignGAN}}

\usepackage{enumitem,xcolor}
\newlist{coloritemize}{itemize}{1}
\setlist[coloritemize]{label=\textcolor{itemizecolor}{\textbullet},font=\bfseries\color{itemizecolor}}
\colorlet{itemizecolor}{red}

\usepackage{accents}

\usepackage[accsupp]{axessibility}  
\usepackage[pagebackref,breaklinks,colorlinks]{hyperref}

\usepackage[title]{appendix}

\usepackage[capitalize]{cleveref}
\crefname{section}{Sec.}{Secs.}
\Crefname{section}{Section}{Sections}
\Crefname{table}{Table}{Tables}
\crefname{table}{Tab.}{Tabs.}

\def\cvprPaperID{} 
\def\confName{CVPR}
\def\confYear{2022}

\begin{document}

\title{Signing at Scale: Learning to Co-Articulate Signs \\ for Large-Scale Photo-Realistic Sign Language Production}

\author{Ben Saunders, Necati Cihan Camgoz, Richard Bowden\\
University of Surrey\\
{\tt\small \{b.saunders, n.camgoz, r.bowden\}@surrey.ac.uk}
}

\twocolumn[{%
\renewcommand\twocolumn[1][]{#1}%
\maketitle
\begin{center}
    \centering
    \includegraphics[width=.99\textwidth]{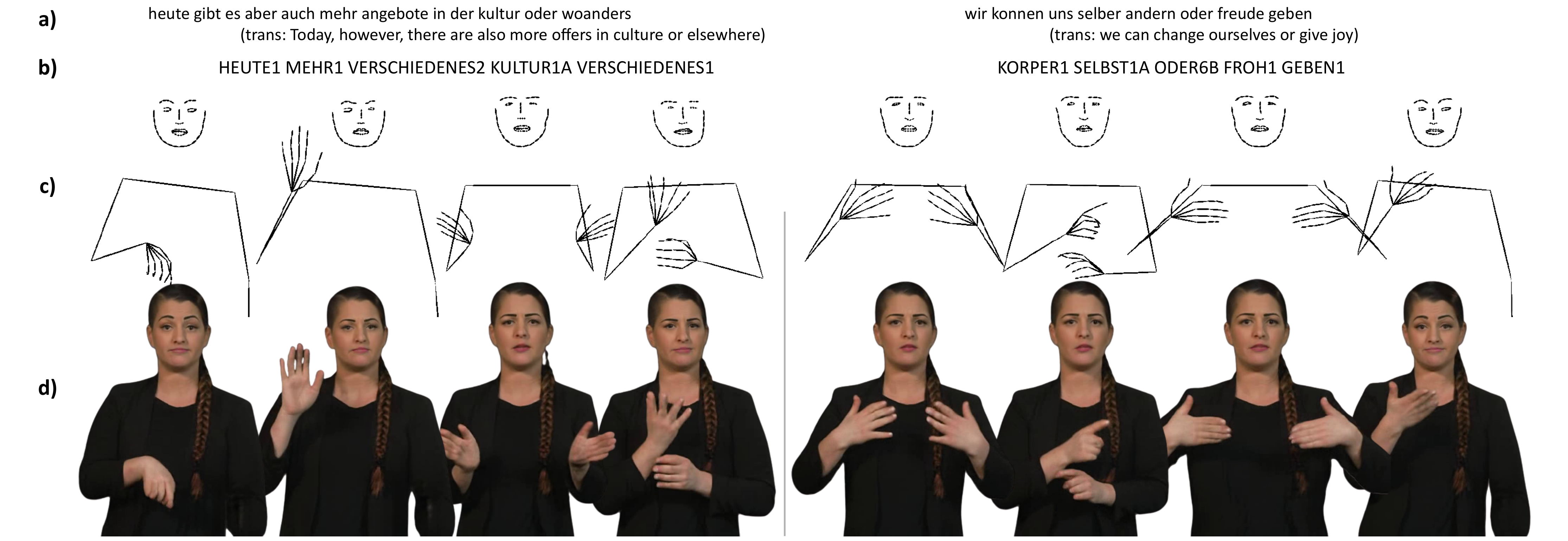}
    \captionof{figure}{\textbf{Photo-Realistic Sign Language Production:} Given a spoken language sentence from an unconstrained domain of discourse (a), an initial translation is conducted to a gloss sequence (b). \FrameSelectionNet{} next produces a co-articulated continuous skeleton pose sequence from dictionary signs (c), which \methodName{} generates into a photo-realistic sign language video in a given style (d).}
    \label{fig:intro}
\end{center}%
}]




\begin{abstract}
   Sign languages are visual languages, with vocabularies as rich as their spoken language counterparts. However, current deep-learning based \acf{slp} models produce under-articulated skeleton pose sequences from constrained vocabularies and this limits applicability. To be understandable and accepted by the deaf, an automatic \ac{slp} system must be able to generate co-articulated photo-realistic signing sequences for large domains of discourse.
   
   In this work, we tackle large-scale \ac{slp} by learning to co-articulate between dictionary signs, a method capable of producing smooth signing while scaling to unconstrained domains of discourse. To learn sign co-articulation, we propose a novel \FrameSelectionNetLong{} (\FrameSelectionNet{}) that improves the temporal alignment of interpolated dictionary signs to continuous signing sequences. Additionally, we propose \methodName{}, a pose-conditioned human synthesis model that produces photo-realistic sign language videos direct from skeleton pose. We propose a novel keypoint-based loss function which improves the quality of synthesized hand images. 
   
   We evaluate our \ac{slp} model on the large-scale \ac{mdgs} corpus, conducting extensive user evaluation showing our \FrameSelectionNet{} approach improves co-articulation of interpolated dictionary signs. Additionally, we show that \methodName{} significantly outperforms all baseline methods for quantitative metrics, human perceptual studies and native deaf signer comprehension.
\end{abstract}

\section{Introduction} \label{sec:intro}

Sign languages are rich visual languages with large lexical vocabularies \cite{stokoe1980sign} and intricate co-articulated movements of both manual (hands and body) and non-manual (facial) features. \acf{slp}, the automatic translation from spoken language sentences to sign language sequences, must be able to produce photo-realistic continuous signing for large domains of discourse to be useful to the deaf communities.

Prior deep-learning approaches to \ac{slp} have either produced concatenated isolated sequences that disregard the natural co-articulation between signs \cite{stoll2018sign,zelinka2020neural} or continuous sequences end-to-end \cite{saunders2021continuous,saunders2020progressive,zelinka2020neural,huang2021towards} which suffer from under-articulation \cite{saunders2020adversarial}. Furthermore, these methods have struggled to generalise beyond the limited domain of weather \cite{forster2012rwth}.

In this paper, we propose an \ac{slp} method to produce photo-realistic continuous sign language videos direct from unconstrained spoken language sequences. Firstly, we translate from spoken language to gloss\footnote{Glosses are a written representation of sign that follow sign language ordering and grammar, defined as minimal lexical items \cite{stokoe1980sign}.} sequences. We next learn the temporal co-articulation between gloss-based dictionary signs, modelling the temporal prosody of sign language \cite{brentari2018production}.

\begin{figure*}[t!]
    \centering
    \includegraphics[width=0.9\linewidth]{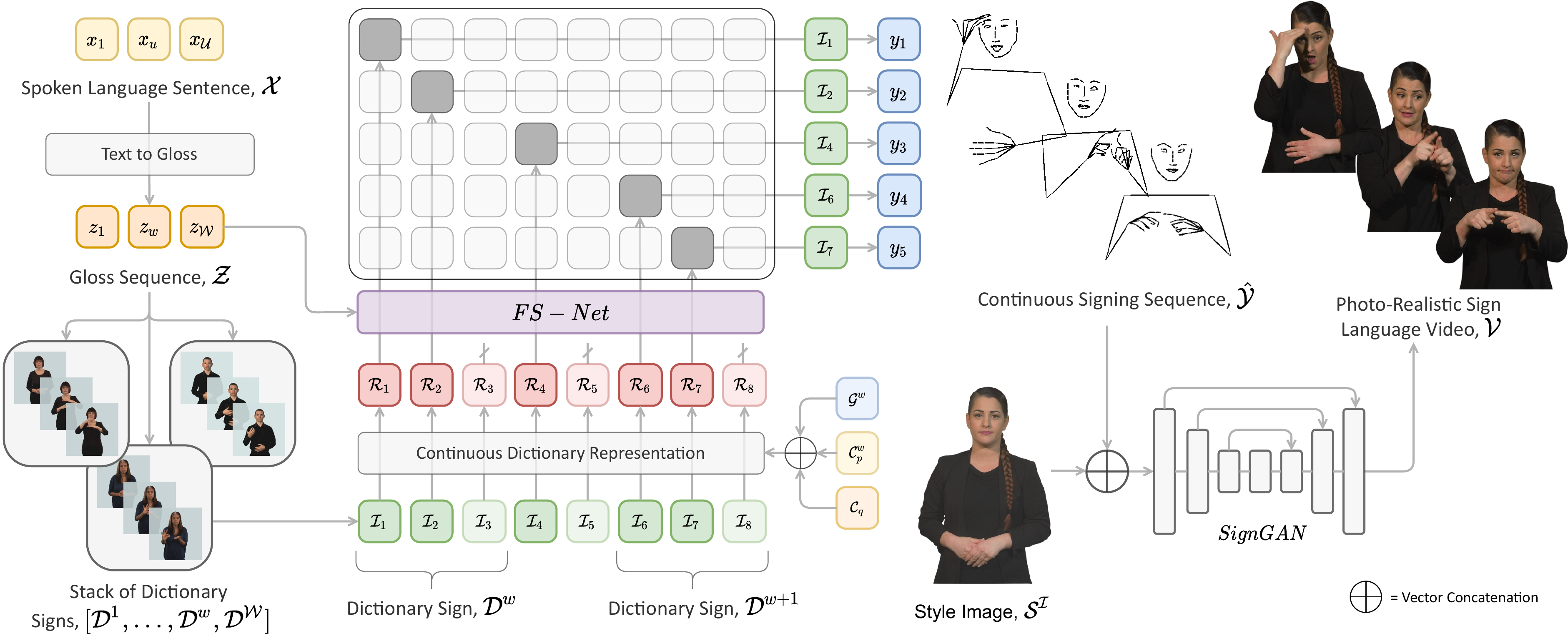}
    \caption{Overview of our proposed large-scale \ac{slp} method. An initial Text to Gloss (left) animates an interpolated dictionary sequence, $\mathcal{I}$, with a Frame Selection Network (\FrameSelectionNet{}), learning the temporal alignment (middle) to a continuous signing sequence, $\mathcal{Y}$. Finally, \methodName{} generates a photo-realistic sign language video, $\mathcal{V}$, (right) from the continuous skeleton pose and a given style image, $\mathcal{S}^{\mathcal{I}}$.}
    \label{fig:Model_Overview}
\end{figure*}

To model sign co-articulation, we propose a novel \FrameSelectionNetLong{} (\FrameSelectionNet{}) that learns the optimal subset of frames that best represents a continuous signing sequence (Fig. \ref{fig:Model_Overview} middle). We build a transformer encoder with cross-attention \cite{vaswani2017attention} to predict a temporal alignment path supervised by \ac{dtw}.

The resulting skeleton pose sequences are subsequently used to condition a video-to-video synthesis model capable of generating photo-realistic sign language videos, named \methodName{} (Fig. \ref{fig:Model_Overview} right). Due to the natural presence of motion blur in sign language datasets from fast moving hands \cite{forster2014extensions}, a classical application of a hand discriminator leads to an increase in blurred hand generation. To avoid this, we propose a novel keypoint-based loss that significantly improves the quality of hand image synthesis in our photo-realistic signer generation module. To enable training on diverse sign language datasets, we propose a method for controllable video generation that models a multi-modal distribution of sign language videos in different styles.

Our deep-learning based \ac{slp} model is able to generalise to large domains of discourse, as it is trivial to increase vocabulary with a few examples of this new sign in a continuous signing context. We conduct extensive deaf user evaluation on a translation protocol of \ac{mdgs} \cite{hanke2010dgs}, showing that \FrameSelectionNet{} improves the natural signing motion of interpolated dictionary sequences and is overwhelmingly preferred to baseline \ac{slp} methods \cite{saunders2021mixed}. Additionally, we achieve state-of-the-art back translation performance on \ac{ph14t} with a 43\% improvement over baselines, highlighting the understandable nature of our approach.

Furthermore, we evaluate \methodName{} using the high quality \ac{c4a} dataset \cite{camgoz2021content4all}, outperforming state-of-the-art synthesis methods \cite{chan2019everybody,stoll2020text2sign,wang2018video,wang2018high} for quantitative evaluation and human perception studies. Finally, we conduct a further deaf user evaluation to show that \methodName{} is more understandable than the skeletal sequences previously used to represent sign \cite{saunders2020progressive}.

\noindent The contributions of this paper can be summarised as:
\begin{itemize}
    \item The first SLP model to produce large-scale sign language sequences from an unconstrained domain of discourse to a level understandable by a native deaf signer
    \item A novel Frame Selection Network, \FrameSelectionNet, that learns to co-articulate between dictionary signs via a monotonic alignment to continuous sequences
    \item A method to generate photo-realistic continuous sign language videos, \methodName{}, with a novel hand keypoint loss that improves the hand synthesis quality
    \item Extensive user evaluation of our proposed approach, showing preference of our proposed method, alongside state-of-the-art back translation results
\end{itemize}

\section{Related Work} \label{sec:related_work}
\paragraph{Sign Language Production}
The initial focus of computational sign language technology was \ac{slr} \cite{cui2017recurrent,grobel1997isolated,koller2020quantitative} with few works tackling unconstrained \ac{slr} \cite{kadir2004minimal,cooper2007large,koller2015continuous}. More recently, focus has shifted to \ac{slt} \cite{camgoz2018neural,camgoz2020sign,ko2019neural}.

\acf{slp}, the translation from spoken to sign language, has been historically tackled using animated avatars \cite{cox2002tessa,karpouzis2007educational,mcdonald2016automated} with rules-based co-articulation that does not generalise to unseen sequences \cite{segouat2009study}. 

Initial deep learning-based \ac{slp} methods concatenated isolated signs with no regards for natural co-articulation \cite{stoll2018sign,zelinka2020neural}. Recently, continuous \ac{slp} methods have directly regressed sequences of multiple signs \cite{saunders2020progressive,saunders2021continuous,huang2021towards,saunders2021mixed,saunders2021skeletal}, but exhibit under-articulated signing motion due to regression to the mean. To overcome under-articulation, we avoid generating pose directly and learn the optimal temporal alignment between dictionary and continuous sign sequences. 

In addition, prior work has represented sign language as skeleton pose sequences, which have been shown to reduce the deaf comprehension compared to a photo-realistic production \cite{ventura2020can}. Previous works have attempted photo-realistic signer generation \cite{stoll2020text2sign,cui2019deep,saunders2021anonysign}, but of low-resolution isolated signs. In this work, we produce high-resolution photo-realistic continuous sign language videos directly from spoken language input, from unrestricted domains of discourse. 

\paragraph{Pose-Conditioned Human Synthesis}

\acfp{gan} \cite{goodfellow2014generative} have achieved impressive results in image \cite{isola2017image,radford2015unsupervised,wang2018high,zhu2017unpaired} and, more recently, video generation tasks \cite{mallya2020world,tulyakov2018mocogan,vondrick2016generating,wang2019few,wang2018video}. Specific to pose-conditioned human synthesis, there has been concurrent research focusing on the generation of whole body \cite{balakrishnan2018synthesizing,ma2017pose,men2020controllable,siarohin2018deformable,tang2020xinggan,zhu2019progressive}, face \cite{deng2020disentangled,kowalski2020config,zakharov2019few} and hand \cite{liu2019gesture,tang2018gesturegan,wu2020mm} images.

However, there has been no research into accurate hand generation in the context of full body synthesis, with current methods failing to generate high-quality hand images \cite{ventura2020can}. Due to the hands being high fidelity objects, they are often overlooked in model optimisation. Chan \etal introduced FaceGAN for high resolution face generation \cite{chan2019everybody}, but no similar work has been proposed for the more challenging task of hand synthesis in the context of sign language, where hand to hand interaction is ubiquitous. In this work, we propose a keypoint-based loss to enhance hand synthesis.

The task of human motion transfer, transferring motion from source to target videos via keypoint extraction, is relevant to our task \cite{chan2019everybody,wei2020gac,zhou2019dance}. However, there has been limited research into the generation of novel poses, which we produce from a given spoken language sentence. Additionally, works have attempted to produce unseen appearances in a few-shot manner \cite{wang2019few,zakharov2019few}, but continue to produce only a single style at inference. 

\paragraph{Sign Language Co-Articulation}

Sign language co-articulation can be defined as ``the articulatory influence of one phonetic element on another across more than one intervening element" \cite{grosvald2009long} and is an important distinction between isolated and natural continuous signing \cite{naert2017coarticulation}.

Co-articulation involves both the motion and duration of signs, with a particular focus on the transition between signs \cite{naert2017coarticulation}. The boundaries of a sign are also modified depending on the context, with continuous signing typically produced faster than their isolated counterparts \cite{segouat2009study}. In this work, we model temporal co-articulation by learning the optimal alignment between isolated signs and continuous signing sequences, predicting the duration, boundary and transition of each sign in context. 

\section{Large-Scale Photo-Realistic \ac{slp}} \label{sec:methodology}

The true aim of a large-scale \ac{slp} model is to translate a spoken language sequence from an unconstrained domain of discourse, \hbox{$\mathcal{X} = (x_{1},...,x_{\mathcal{U}})$} with $\mathcal{U}$ words, to a continuous photo-realistic sign language video, \hbox{$\mathcal{V}^{\mathcal{S}} = (v_{1},...,v_{\mathcal{T}})$} with $\mathcal{T}$ frame. This is a challenging task due to the large vocabulary of unconstrained signing and the intricate spatial nature of sign, with a requirement for temporal co-articulation indicative of natural continuous signing. 

We approach this problem as a multi-stage sequence-to-sequence task. Firstly, spoken language is translated to sign gloss, \hbox{$\mathcal{Z} = (z_{1},...,z_{\mathcal{W}})$}, as an intermediate representation (Sec. \ref{sec:text_to_gloss}). Next, our \FrameSelectionNet{} model co-articulates between gloss-based dictionary signs to produce a full continuous signing sequence, \hbox{$\mathcal{Y} = (y_{1},...,y_{\mathcal{T}})$} (Sec. \ref{sec:gloss_to_sign}). Finally, given $\mathcal{Y}$ and a style image, $\mathcal{S}^{\mathcal{I}}$, our video-to-video signer generation module generates a photo-realistic sign language video, $\mathcal{Z}^{\mathcal{S}}$ (Sec. \ref{sec:sign_to_video}). An overview of our approach can be seen in Fig. \ref{fig:Model_Overview}. In the remainder of this section, we shall describe each component of our approach in detail.

\subsection{Text to Gloss} \label{sec:text_to_gloss}

Given a spoken language sequence, $\mathcal{X}$, we first translate to a sign language grammar and order, represented by a gloss sequence, \hbox{$\mathcal{Z} = (z_{1},...,z_{\mathcal{W}})$} with $\mathcal{W}$ glosses (Fig. \ref{fig:Model_Overview} left). We formulate this as a sequence-to-sequence problem, due to the non-monotonic relationship between the two sequences of different lengths. We use an encoder-decoder transformer \cite{vaswani2017attention} to perform this translation, formalised as: 
\begin{equation}
\label{eq:T2G_encoder}
    f_{t} = E_{T2G}(x_{t} | x_{1:\mathcal{T}})
\end{equation}
\begin{equation}
\label{eq:T2G_decoder}
    g_{w+1} = D_{T2G}(g_{w}  | g_{1:w-1} , f_{1:\mathcal{T}})
\end{equation}
where $f_{t}$ and $g_{w}$ are the encoded source and target tokens respectively and $g_{0}$ is the encoding of the special $\mathrm{<bos>}$ token. The output gloss tokens can be computed as $z_{w} = \operatorname*{argmax}_{i} (g_{w})$ until the special $\mathrm{<eos>}$ token is predicted.

\subsection{Gloss to Pose} \label{sec:gloss_to_sign}

Next, motivated by the monotonic relationship between glosses and signs, we produce a continuous signing pose sequence, $\hat{\mathcal{Y}} = (y_{1},...,y_{\mathcal{T}})$ with $\mathcal{T}$ frames, from the translated gloss sequence, $\mathcal{Z}$, using a learnt co-articulation of dictionary signs. We first encode the gloss sequence using a transformer encoder with self-attention, as:
\begin{equation}
    h_{w} = E_{G2S}(z_{w}  | z_{1:\mathcal{W}})
\end{equation}
where $h_{w}$ is the encoded gloss token for step $w$. We next collect a dictionary sample, $\mathcal{D}^{w}$, for every sign present in the gloss vocabulary. By definition, dictionary signs contain accurate and articulated sign content. Furthermore, it is trivial to expand to larger domains of discourse, simply collecting dictionary examples of the expanded vocabulary.
\paragraph{Interpolated Dictionary Representation}

Given the translated gloss sequence, $\mathcal{Z}$, we create a stack of ordered dictionary signs, $[\mathcal{D}^{1},...,\mathcal{D}^{w},\mathcal{D}^{\mathcal{W}}]$ (Bottom left of Fig. \ref{fig:Model_Overview}). As in previous works \cite{saunders2021continuous}, we represent each dictionary sign as a sequence of skeleton pose, $\mathcal{D}^{w} = (s^{w}_{1},...,s^{w}_{\mathcal{P}^{w}})$ with $\mathcal{P}^{w}$ frames. We first convert the stack of dictionary signs into a continuous sequence by linearly interpolating between neighbouring signs for a predefined fixed $\mathcal{N}_{LI}$ frames. The final interpolated dictionary sequence, $\mathcal{I} = (\mathcal{I}_{1},...,\mathcal{I}_{\mathcal{Q}})$ with $\mathcal{Q}$ frames, is the combination of skeleton pose and the respective linear interpolation.

We next build a continuous dictionary sequence representation to be used as input to \FrameSelectionNet{}. Alongside the skeleton pose of $\mathcal{I}$, we learn a gloss embedding, $\mathcal{G}^{w}$, unique to each gloss in the vocabulary, with a separate shared embedding for all interpolation frames, $\mathcal{G}^{LI}$. Additionally, we use a counter embedding proposed by Saunders \etal \cite{saunders2020progressive}, expanded to both a specific counter, $\mathcal{C}^{w}_{p}$, relating to the progression of each dictionary sign and a global counter, $\mathcal{C}_{q}$, relating to the progress of the full sequence, $\mathcal{I}$. The final continuous dictionary representation, $\mathcal{R} = (\mathcal{R}_{1},...,\mathcal{R}_{\mathcal{Q}})$ with $\mathcal{Q}$ frames, is constructed by concatenating the corresponding skeleton, gloss and counter embeddings per frame, as:
\begin{equation}
    \mathcal{R}_{q} = [s^{w}_{p},\mathcal{G}^{w},\mathcal{C}^{w}_{p},\mathcal{C}_{q}]
\label{eq:sign_representation}
\end{equation}
where frame $q$ represents a time step $p$ frames into gloss $w$. 

\paragraph{Frame Selection Network}

To co-articulate between dictionary signs, we propose a \FrameSelectionNetLong{} (\FrameSelectionNet{}) that learns to predict the temporal alignment to a continuous signing sequence, $\mathcal{Y} = (y_{0},...,y_{\mathcal{T}})$ with $\mathcal{T}$ frames (Fig. \ref{fig:Model_Overview} middle). We note that this is a monotonic sequence-to-sequence task, due to the matching order of signing and the different sequence lengths ($\mathcal{Q} \neq \mathcal{T}$).

Formally, \FrameSelectionNet{} predicts a discrete sparse monotonic temporal alignment path, \hbox{$\hat{\mathcal{A}} \in \mathbb{R}^{\mathcal{Q} \times \mathcal{Q}}$}:
\begin{equation}
    \hat{\mathcal{A}}  = \FrameSelectionNet{}(\mathcal{R}, h_{1:\mathcal{W}})
\end{equation}
where $\hat{\mathcal{A}}$ contains binary decisions representing either frame selection or skipping. Fig. \ref{fig:Model_Overview} shows an example alignment that skips the production of frames $3,5$ and $8$ in the output sequence, removing redundant frames to create a smoother co-articulated continuous sequence. We build \FrameSelectionNet{} as a transformer encoder \cite{vaswani2017attention} with an additional cross-attention to the encoded gloss sequence. To produce the co-articulated continuous signing pose sequence, $\hat{\mathcal{Y}}$, a matrix multiplication can be applied between $\mathcal{I}$ and $\hat{\mathcal{A}}$, as:

\begin{equation}
    \label{eq:continuous_output}
    \hat{\mathcal{Y}} = \mathcal{I} \times \hat{\mathcal{A}}
\end{equation}

This enables the mapping between varied length sequences, with the end of sequence prediction determined as the alignment selection of the final dictionary frame.

\paragraph{Dynamic Time Warping Supervision}

In practice, directly predicting the 2D alignment, $\hat{\mathcal{A}}$, provides weak gradients due to the sparse nature of the alignment. We therefore propose to train \FrameSelectionNet{} using a \acf{dtw} supervision signal \cite{berndt1994dtw} designed to learn the optimal monotonic temporal alignment. We pre-compute the \ac{dtw} path, $\mathcal{A}^{*} = \textsc{DTW}(\mathcal{Q},\mathcal{T})$, between the interpolated dictionary sequence, $\mathcal{I}$, and the target continuous sequence, $\mathcal{Y}$. Due to the intractability of 2D alignment path prediction, we collapse the alignment down to a 1D sequence during training, \hbox{$\hat{\textrm{A}} \in \mathbb{R}^{\mathcal{Q}} = \argmax_q ( \hat{\mathcal{A}})$}. This enables a temporal mask prediction over $\mathcal{I}$, selecting which frames of the interpolated dictionary sequence to animate in turn to create a continuous sequence.

We argue that for the majority of sequences (88\% for \ac{mdgs}), \hbox{$\mathcal{Q} >> \mathcal{T}$}, due to the faster tempo of continuous sign \cite{naert2017coarticulation}. We thus assume that no frames are added during temporal alignment, only removed. To train \FrameSelectionNet{}, we compute a cross entropy loss $\mathcal{L}_{CE}$ between the predicted 1D temporal alignment, \hbox{$\hat{\textrm{A}} \in \mathbb{R}^{\mathcal{Q}}$}, and the ground truth \ac{dtw} alignment, $\textrm{A}^{*} \in \mathbb{R}^{\mathcal{Q} \times 1}$, as: 
\begin{equation}
    \mathcal{L}_{CE} (\hat{\textrm{A}},\textrm{A}^{*}) = - \frac{1}{\mathcal{Q}} \sum_{q=1}^{\mathcal{Q}} \textrm{A}^{*}_{q} \cdot{} \log (\hat{\textrm{A}}_{q})
\end{equation}

The final continuous sign pose sequence, \hbox{$\hat{\mathcal{Y}} = (y_{1},...,y_{\mathcal{T}})$}, is produced as shown in Eq. \ref{eq:continuous_output}.

\subsection{Pose to Video} \label{sec:sign_to_video}

To generate a photo-realistic sign language video, $\mathcal{V}^{\mathcal{S}}$, conditioned on the produced sign pose sequence, $\hat{\mathcal{Y}}$, we propose a method for video-to-video signer generation, \methodName{} (Fig. \ref{fig:Model_Overview} right). Taking inspiration from \cite{chan2019everybody}, in the conditional \ac{gan} setup, a generator network, $G$, competes in a min-max game against a multi-scale discriminator, $D = (D_{1}, D_{2}, D_{3})$. The goal of $G$ is to synthesise images of similar quality to ground-truth images, in order to fool $D$. Conversely, the aim of $D$ is to discern the ``fake'' images from the ``real'' images. For our purposes, $G$ synthesises images of a signer, $v^{\mathcal{S}}$ given a human pose, $y_{t}$, and a style image, $\mathcal{S}^{\mathcal{I}}$.

Following \cite{isola2017image}, we introduce skip connections to the architecture of $G$ in a \textit{U-Net} structure \cite{ronneberger2015u} between each down-sampling layer $i$ and up-sampling layer $n-i$, where $n$ is the total number of up-sampling layers. Skip connections propagate pose information across the networks, enabling the generation of fine-grained details. Specifically, we add skip connections between each down-sampling layer $i$ and up-sampling layer $n-i$, where $n$ is the total number of up-sampling layers. 

\paragraph{Controllable Video Generation} 

To enable training on diverse sign language datasets, we use a style-controllable video generation approach \cite{saunders2021anonysign}. A style image, $\mathcal{S}^{\mathcal{I}}$, is provided to condition synthesis alongside the pose sequence, as seen in Figure \ref{fig:Model_Overview}. \methodName{} learns to associate the given style, $\mathcal{S}$, with the person-specific aspects of the corresponding target image, $v^{\mathcal{S}}_{t}$, such as the clothing or face, but disentangle the signer-invariant skeleton pose. 

\begin{figure*}[t!]
    \centering
    \includegraphics[width=0.8\linewidth]{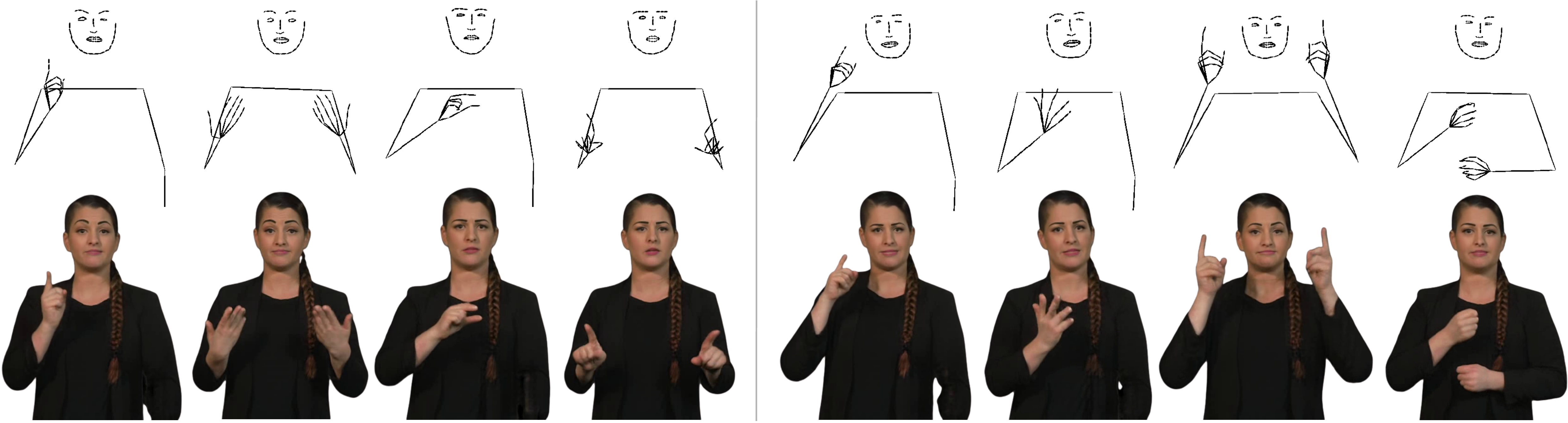}
    \caption{Example photo-realistic frames with skeleton pose produced using \FrameSelectionNet{} and photo-realistic video generated using \methodName{}.}
    \label{fig:Qual_Eval}
    \vspace{-0.4cm}
\end{figure*}

Controllable generation allows \methodName{} to make use of the variability in signer appearance in the data. A multi-modal distribution of sign language videos in different styles, $\mathcal{V}^{S}$, can be produced, where \hbox{$S \in \{1,N_{S}\}$} represents the styles seen during training \footnote{For qualitative examples (e.g. in Fig \ref{fig:Qual_Eval}), we share a single signer appearance, as we have consent from this signer to use their appearance for publication purposes.}.

\begin{figure}[t!]
    \centering
    \includegraphics[width=0.9\linewidth]{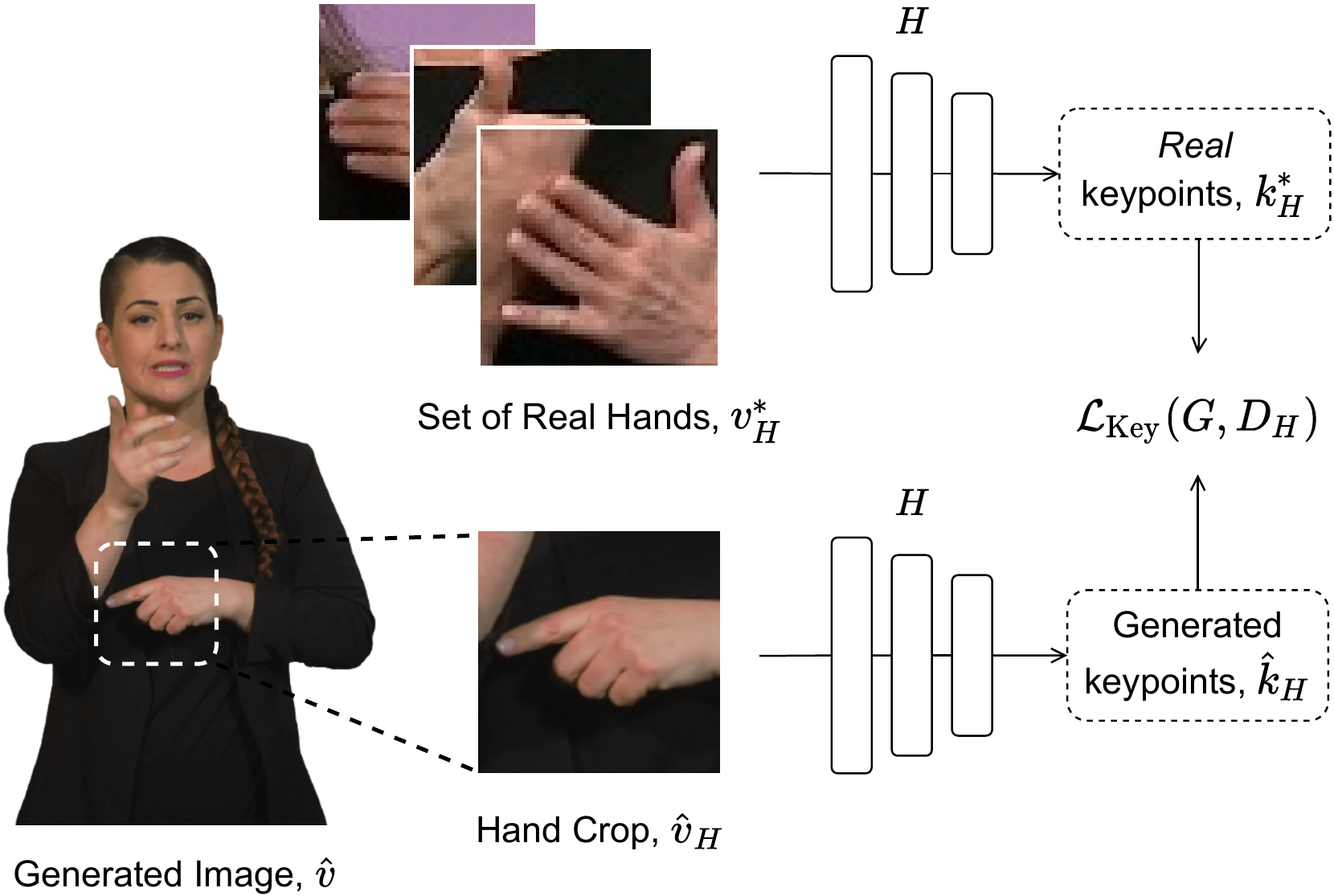}
    \caption{Hand keypoint loss overview. A keypoint discriminator, $D_{H}$, compares keypoints extracted from generated hands, $\hat{k}_{H}$, and real hands, ${k}^{*}_{H}$.}
    \label{fig:keypoint_loss}
\end{figure}%

\paragraph{Hand Keypoint Loss}

Previous pose-conditioned human synthesis methods have failed to generate realistic and accurate hand images \cite{ventura2020can}. To enhance the quality of hand synthesis, we introduce a novel loss that operates in the keypoint space, as shown in Figure \ref{fig:keypoint_loss}. A pre-trained 2D hand pose estimator \cite{ge20193d}, $H$, is used to extract hand keypoints, $k_{H}$, from cropped hand regions (\textit{i.e.} a 60x60 patch centered around the middle knuckle), $v_{H}$, as \hbox{$k_{H} = H(v_{H})$}. We avoid operating in the image space due to the existence of blurry hand images in the dataset, whereas the extracted keypoints are invariant to motion-blur. A hand keypoint discriminator, $D_{H}$, then attempts to discern between the ``real'' keypoints, \hbox{$k^{\star}_{H} = H(v_{H})$}, and the ``fake'' keypoints, \hbox{$\hat{k}_{H} = H(G(y_{H}))$}, leading to the objective:
\begin{align}
\label{eq:loss_HandKey}
    \mathcal{L}_{\mathrm{\textsc{Key}}}(G,D_{H}) & =  \mathbb{E}_{y_{H},z_{H}} [\log D_{H}(k^{\star}_{H})] \nonumber \\
    + \mathbb{E}_{y_{H}} & [\log (1-D_{H}(\hat{k}_{H}))]
\end{align}

\paragraph{Full Objective}

In standard image-to-image translation frameworks \cite{isola2017image,wang2018high}, $G$ is trained using a combination of adversarial and perceptual losses. We update the multi-scale adversarial loss, $\mathcal{L}_{GAN}(G,D)$, to reflect the controllable generation with a joint conditioning on sign pose, $y_{t}$, and style image, $\mathcal{S}^{\mathcal{I}}$:
\begin{align}
\label{eq:loss_gan}
    \mathcal{L}_{GAN}(G,D) & = \sum^{k}_{i=1} \;  \mathbb{E}_{y_{t},z_{t}} [\log D_{i}(z_{t} \mid y_{t},\mathcal{S}^{\mathcal{I}})] \nonumber \\
    + \mathbb{E}_{y_{t}} [\log & (1-D_{i}(G(y_{t},\mathcal{S}^{\mathcal{I}}) \mid y_{t},\mathcal{S}^{\mathcal{I}}))]  
\end{align}
where $k = 3$ reflects the multi-scale discriminator. The adversarial loss is supplemented with two feature-matching losses; $\mathcal{L}_{FM}(G,D)$, the discriminator feature-matching loss presented in pix2pixHD \cite{wang2018high}, and $\mathcal{L}_{VGG}(G,D)$, the perceptual reconstruction loss \cite{johnson2016perceptual} which compares pretrained VGGNet \cite{simonyan2014very} features at multiple layers of the network. Our full \methodName{} objective, $\mathcal{L}_{Total}$, is a weighted sum of these, alongside our proposed hand keypoint loss (Eq. \ref{eq:loss_HandKey}), as:

\begin{align}
\label{eq:loss_FM}
    \mathcal{L}_{Total} & = 
    \min_{G} (( \max_{D_{i}}  \sum^{k}_{i=1} \; \mathcal{L}_{GAN}(G,D_{i}))  \nonumber  \\
     + \lambda_{FM} & \sum^{k}_{i=1} \; \mathcal{L}_{FM}(G,D_{i}) 
     + \lambda_{VGG} \; \mathcal{L}_{VGG} (G(y_{t},I^{S}),z_{t}) \nonumber  \\
     + \lambda_{\textsc{Key}} & \; \mathcal{L}_{\mathrm{\textsc{Key}}}(G,D_{H}))
\end{align}
\noindent where $k = 3$ and $\lambda_{FM},\lambda_{VGG},\lambda_{\textsc{Key}}$ weight the contributions of each loss.

\section{Experiments} \label{sec:experiments}

In this section, we evaluate our large-scale photo-realistic \ac{slp} approach. We outline our experimental setup then perform quantitative, qualitative \& user evaluation.

\subsection{Experimental Setup}

To train our large-scale \ac{slp} approach, we set a new translation protocol on the Meine DGS (\ac{mdgs}) corpus\footnote{With permission from the University of Hamburg.} \cite{hanke2010dgs}, a large \ac{dgs} linguistic resource capturing free-form signing from 330 deaf participants, with a vocabulary of 10,042 glosses. To adapt the corpus for translation, we segment the free-flowing discourse into 40,230 segments of German sentences, sign gloss translations and sign language videos. We pre-process the \ac{mdgs} gloss annotations \cite{konrad2018public} and create two protocols, with either gloss variants included (\acs{mdgsv}) or removed (\ac{mdgs}). We publicly release these translation protocols\footnote{https://github.com/BenSaunders27/meineDGS-Translation-Protocols} to facilitate the future growth in large-scale \ac{slp} and \ac{slt} research, with further details provided in the appendix. A license must be obtained from the University of Hamburg to use \ac{mdgs} for computational research.

For additional experiments, we use the benchmark \ac{ph14t} dataset \cite{camgoz2018neural} from the constrained weather broadcast domain, with setup and skeletal pose configuration as in \cite{saunders2020progressive}. We collect exhaustive dictionary examples of every \ac{dgs} sign present in \ac{mdgs} and \ac{ph14t}, trimmed to remove the sign onset and offset. For samples without expressive mouthings, we insert the facial features present in a example of the respective gloss from the continuous training set. For photo-realistic video generation, we use the \ac{c4a} dataset \cite{camgoz2021content4all} due to its high video quality and diverse interpreter appearance. We use a heat-map representation as pose condition, with each skeletal limb plotted on a separate feature channel. 

We build our Text to Gloss models with 2 layers, 4 heads and hidden size of 128 and our \FrameSelectionNet{} with 2 layers, 4 heads and 64 hidden size. We set the interpolation frames, $\mathcal{N}_{LI}$, to 5 and the learning rate to $10^{-3}$. Our code is based on JoeyNMT \cite{JoeyNMT}, and implemented using PyTorch \cite{paszke2017automatic}.

\subsection{Quantitative Evaluation} \label{sec:quant_experiments}

\subsubsection{Text to Gloss} \label{sec:Text_to_Gloss_Results}

We first evaluate our Text to Gloss translation described in Sec. \ref{sec:text_to_gloss}. Table \ref{tab:SLT_Baselines} shows a performance of 21.93 BLEU-4 on \ac{ph14t}, outperforming \cite{saunders2020progressive} (20.23) but falling short of \cite{moryossef2021data} (23.17) who use larger training data. Translation performance is considerably lower on both \ac{mdgsv} and \ac{mdgs} due to the larger domain, showing that further research is required to scale the task to larger vocabularies.

\begin{table}[!h]
\centering
\resizebox{0.75\linewidth}{!}{%
\begin{tabular}{@{}p{0.0cm}cc|cc@{}}
\toprule
  & \multicolumn{2}{c}{DEV SET}  & \multicolumn{2}{c}{TEST SET} \\ 
\multicolumn{1}{c}{Dataset:}  & BLEU-4   & \multicolumn{1}{c}{ROUGE}  & BLEU-4  & ROUGE  \\ \midrule
\multicolumn{1}{r|}{\ac{mdgsv}} & 1.96 & 24.51 & 1.16 & 25.34 \\ 
\multicolumn{1}{r|}{\ac{mdgs}} & 3.17 & 32.93 & 3.08 & 32.52 \\ 
\multicolumn{1}{r|}{\ac{ph14t}} & {\B 21.93} & {\B 57.25} & {\B 20.08}  &  {\B 56.63} \\ 
\bottomrule
\end{tabular}%
}
\caption{Text to Gloss results on \ac{mdgs} and \ac{ph14t}.}
\label{tab:SLT_Baselines}
\vspace{-0.6cm}
\end{table}

\begin{table}[!b]
\centering
\resizebox{0.9\linewidth}{!}{%
\begin{tabular}{@{}p{2.8cm}cc|cc@{}}
\toprule
     & \multicolumn{2}{c}{DEV SET}  & \multicolumn{2}{c}{TEST SET} \\ 
\multicolumn{1}{c}{Approach:}  & BLEU-4  & ROUGE & BLEU-4 & ROUGE \\ \midrule
\multicolumn{1}{r|}{Progressive Transformers \cite{saunders2020progressive}} & 11.93 & 34.01 & 10.43 & 32.02 \\
\multicolumn{1}{r|}{Adversarial Training \cite{saunders2020adversarial}} & 13.16 & 36.75 & 12.16 & 34.19 \\
\multicolumn{1}{r|}{Mixture Density Networks \cite{saunders2021continuous}} & 13.14 & 39.06 & 11.94 & 35.19 \\
\multicolumn{1}{r|}{Mixture of Motion Primitives \cite{saunders2021mixed}} & 13.32 & 37.58 & 12.67  & 35.61 \\
\multicolumn{1}{r|}{Interpolated Dictionary Sequence} & 16.28 & 38.11 & 16.27 & 36.95 \\
\multicolumn{1}{r|}{\textbf{\FrameSelectionNet{} (Ours})} & \textbf{19.14} & \textbf{40.94} & \textbf{18.78} & \textbf{40.60} \\
\bottomrule
\end{tabular}%
}
\caption{Back translation results on the \ac{ph14t} dataset for the \textit{Gloss to Pose} task.}
\label{tab:gloss_to_pose}
\end{table}

\subsubsection{Gloss to Pose}

\paragraph{Back Translation}
Back translation has developed as the state-of-the-art \ac{slp} evaluation metric \cite{saunders2020progressive}. We train an \ac{slt} model \cite{camgoz2020sign} on \ac{ph14t} with skeleton pose sequences generated using our \FrameSelectionNet{} approach. Table \ref{tab:gloss_to_pose} shows considerable performance gains (43\%) compared to baseline methods on the Gloss to Pose task \cite{saunders2020progressive,saunders2020adversarial,saunders2021continuous,saunders2021mixed}. This highlights the increased comprehension provided by \FrameSelectionNet{} compared to baseline end-to-end regression methods, with an ability to overcome the poor quality of the \ac{ph14t} dataset. Furthermore, it can be seen that interpolated dictionary sequences (Comparable to that of \cite{stoll2018sign}) achieve worse back translation results, highlighting the effect of \FrameSelectionNet{} co-articulation for comprehension. 

Additionally, Table \ref{tab:text_to_pose} shows further state-of-the-art results on the full pipeline of text to pose, with an initial text to gloss translation and a subsequent sign animation. This highlights the effectiveness of the full spoken language to photo-realistic video pipeline required for true \ac{slp}.

\begin{table}[!t]
\centering
\resizebox{0.9\linewidth}{!}{%
\begin{tabular}{@{}p{2.8cm}cc|cc@{}}
\toprule
     & \multicolumn{2}{c}{DEV SET}  & \multicolumn{2}{c}{TEST SET} \\ 
\multicolumn{1}{c}{Approach:}  & BLEU-4  & ROUGE & BLEU-4 & ROUGE \\ \midrule
\multicolumn{1}{r|}{Progressive Transformers \cite{saunders2020progressive}} & 11.82 & 33.18 & 10.51 & 32.46 \\ 
\multicolumn{1}{r|}{Adversarial Training \cite{saunders2020adversarial}} & 12.65 & 33.68 & 10.81 & 32.74 \\
\multicolumn{1}{r|}{Mixture Density Networks \cite{saunders2021continuous}} & 11.54 & 33.40 & 11.68 & 33.19 \\ 
\multicolumn{1}{r|}{Mixture of Motion Primitives \cite{saunders2021mixed}} & 14.03 & \textbf{37.76} & 13.30 & 36.77 \\
\multicolumn{1}{r|}{\textbf{\FrameSelectionNet{} (Ours})} & \textbf{16.92}  & 35.74 & \textbf{21.10} & \textbf{42.57} \\
\bottomrule
\end{tabular}%
}
\caption{Back translation results on the \ac{ph14t} dataset for the \textit{Text to Pose} task.}
\label{tab:text_to_pose}
\end{table}

\paragraph{Sign User Evaluation}

We next perform extensive user evaluation with native signers, animating our skeleton pose outputs using \methodName{}. All baselines are also generated by \methodName{}, to alleviate visual differences in comparison. In total, 10 participants completed our sign user evaluation, of which all were fluent signers and 20\% were deaf. We provide all the generated user evaluation videos in the supplementary materials.

We first compare the comprehension of \FrameSelectionNet{} compared to the state-of-the-art deep learning based \ac{slp} method \cite{saunders2021mixed}. We show participants pairs of generated videos from the same sequence, asking to select which video was more understandable. Table \ref{tab:understandable_baselines} shows how our productions were unanimously preferred to baselines for both \ac{mdgs} and \ac{ph14t}. This overwhelming result highlights both the increased comprehension of \FrameSelectionNet{}, alongside the inability of previous methods to scale to unconstrained domains of discourse.

\begin{table}[b!]
\centering
\resizebox{0.70\linewidth}{!}{%
\begin{tabular}{@{}p{2.0cm}|ccc@{}}
\toprule
Dataset & \FrameSelectionNet{} & Baseline \cite{saunders2021mixed} & Equal \\\midrule
\multirow{1}{*}{\ac{mdgs}}  & \textbf{95\%} & 0\% & 5\% \\ \midrule
\multirow{1}{*}{\ac{ph14t}} & \textbf{95\%} & 0\% & 5\% \\

\bottomrule
\end{tabular}%
}
\caption{Comprehension user evaluation results, showing the percentage of participants who chose productions from \FrameSelectionNet{} or a baseline \cite{saunders2021mixed} to be more understandable, or equal.}
\label{tab:understandable_baselines}
\end{table}

We next evaluate how understandable our large-scale sign productions are in isolation. We show each participant a produced video alongside a list of 10 signs, of which 5 are signed in the video, and ask them to select which signs they believe are being signed. For \FrameSelectionNet{} productions, an average of 4.8 signs were recognised for each video. This shows our productions are easily understandable by native signers, an essential result for accurate large-scale \ac{slp}.

Our final user evaluation evaluates how co-articulated our \FrameSelectionNet{} productions are. We show participants two videos of the same sequence, one isolated dictionary sequence and one co-articulated continuous video generated by \FrameSelectionNet{}, and ask them to select which they believed had the most smooth transitions between signs. We first evaluate dictionary signs without trimming the sign onset and offset, with Table \ref{tab:coarticulated_signing} showing our productions were unanimously preferred. Moving to trimmed dictionary signs, \FrameSelectionNet{} productions were preferred 40\% of the time, with 13\% equal preference. This highlights the effectiveness of \FrameSelectionNet{} at improving co-articulation between dictionary signs and temporally aligning to continuous signing sequences.

\begin{table}[t!]
\centering
\resizebox{0.70\linewidth}{!}{%
\begin{tabular}{@{}p{2.5cm}ccc@{}}
\toprule
  & \FrameSelectionNet{} & Isolated & Equal \\\midrule
\multicolumn{1}{r|}{Non-Trimmed} & \textbf{100}\% & 0\% & 0\% \\ 
\multicolumn{1}{r|}{Trimmed} & 40\% & \textbf{47}\% & 13\% \\ 
\bottomrule
\end{tabular}%
}
\caption{Co-articulation user evaluation results, showing the percentage of participants who believed the video with the smoothest transitions was from \FrameSelectionNet{}, isolated concatenation or equal.}
\label{tab:coarticulated_signing}
\end{table}

\begin{table}[b!]
\centering
\resizebox{0.9\linewidth}{!}{%
\begin{tabular}{@{}p{3.0cm}cccc@{}}
\toprule
  & \multicolumn{1}{c}{SSIM $\uparrow$} & \multicolumn{1}{c}{Hand SSIM $\uparrow$} & \multicolumn{1}{c}{Hand Pose $\downarrow$} & \multicolumn{1}{c}{FID $\downarrow$} \\ \midrule
\multicolumn{1}{r|}{EDN \cite{chan2019everybody}} & 0.737 & 0.553 & 23.09 & 41.54 \\
\multicolumn{1}{r|}{vid2vid \cite{wang2018video}} & 0.750  & 0.570 & 22.51 & 56.17 \\
\multicolumn{1}{r|}{Pix2PixHD \cite{wang2018high}} & 0.737 & 0.553 & 23.06 & 42.57 \\ 
\multicolumn{1}{r|}{Stoll \etal \cite{stoll2020text2sign}} & 0.727 & 0.533 & 23.17 & 64.01 \\ 
\multicolumn{1}{r|}{\methodName{} (Ours)} & \textbf{0.759} & \textbf{0.605} & \textbf{22.05} & \textbf{27.75}  \\
\bottomrule
\end{tabular}%
}
\caption{Baseline model comparison results of photo-realistic sign language video generation.}
\label{tab:baselines_diverse}
\end{table}

\subsubsection{Pose to Video}

Finally, we evaluate our photo-realistic sign language video approach, \methodName{}. We compare the performance of \methodName{} against state-of-the-art image-to-image and video-to-video translation methods \cite{chan2019everybody,wang2018video,wang2018high,stoll2020text2sign}, conditioned on skeletal pose images. We measure the quality of synthesized images using the following metrics; 1) SSIM: Structural Similarity \cite{wang2004image} over the full image. 2) Hand SSIM: SSIM metric over a crop of each hand. 3) Hand Pose: Absolute distance between 2D hand keypoints of the produced and ground truth hand images, using a pre-trained hand pose estimation model \cite{ge20193d}. 4) FID: Fr\'echet Inception Distance \cite{heusel2017gans} over the full image.

\paragraph{Baseline Comparison}

We first compare \methodName{} to baseline methods for photo-realistic generation given a sequence of ground truth poses as input. Table \ref{tab:baselines_diverse} shows results on the \ac{c4a} data, with \methodName{} outperforming all baselines particularly for the Hand SSIM and FID. We believe this is due to the improved quality of synthesized hand images by using the proposed hand keypoint loss.

\paragraph{Ablation Study}

We perform an ablation study of \methodName{}, with results in Table \ref{tab:ablation}. As suggested in Sec. \ref{sec:sign_to_video}, the hand discriminator performs poorly for both SSIM and hand SSIM, due to the generation of blurred hands. However, our proposed hand keypoint loss increases model performance considerably and particularly for Hand SSIM, emphasizing the importance of an adversarial loss invariant to blurring.

\begin{table}[t!]
\centering
\resizebox{0.9\linewidth}{!}{%
\begin{tabular}{@{}p{3.0cm}ccccc@{}}
\toprule
  & \multicolumn{1}{c}{SSIM $\uparrow$} & \multicolumn{1}{c}{Hand SSIM $\uparrow$} & \multicolumn{1}{c}{Hand Pose $\downarrow$} & \multicolumn{1}{c}{FID $\downarrow$} \\ \midrule
\multicolumn{1}{r|}{Baseline} & 0.743 & 0.582 & 22.87 & 39.33 \\
\multicolumn{1}{r|}{Hand Discriminator} & 0.738 & 0.565 & 22.81 & 39.22 \\
\multicolumn{1}{r|}{Hand Keypoint Loss} & \textbf{0.759} & \textbf{0.605} & \textbf{22.05} & \textbf{27.75} \\
\bottomrule
\end{tabular}%
}
\caption{Ablation study results of \methodName{}}
\label{tab:ablation}
\vspace{-0.4cm}
\end{table}

\begin{table}[b!]
\centering
\resizebox{0.5\linewidth}{!}{%
\begin{tabular}{@{}p{3.0cm}cc@{}}
\toprule
& \multicolumn{1}{c}{Body} & \multicolumn{1}{c}{Hand}  \\ \midrule
\multicolumn{1}{r|}{EDN \cite{chan2019everybody}} & 100\% & 97.8\% \\
\multicolumn{1}{r|}{vid2vid \cite{wang2018video}} & 85.9\% & 84.8\% \\
\multicolumn{1}{r|}{Pix2PixHD \cite{wang2018high}} & 98.9\% & 100\% \\
\multicolumn{1}{r|}{Stoll \etal \cite{stoll2020text2sign}} & 100\% & 100\% \\
\bottomrule
\end{tabular}%
}
\caption{Perceptual study results, showing the percentage of participants who preferred \methodName{} to the baseline model.}
\label{tab:perceptual}
\end{table}

\paragraph{Perceptual Study}

We perform an additional perceptual study of our photo-realistic generation, showing participants pairs of 10 second videos generated by \methodName{} and a corresponding baseline method.  Participants were asked to select which video was more visually realistic, with a separate focus on the body and hands. In total, 46 participant completed the study, of which 28\% were signers, each viewing 2 randomly selected videos from each baseline. Table \ref{tab:perceptual} shows the percentage of participants who preferred the outputs of \methodName{} to the baseline method. It can be seen that \methodName{} outputs were unanimously preferred for both body (96.2\% average) and hand (95.6\% average) synthesis. Vid2vid \cite{wang2018video} was the strongest contender, with our productions preferred only 85\% of the time.

\paragraph{Deaf User Evaluation}

Our final user evaluation compares the comprehension of photo-realistic videos against the previously-used skeletal pose representation \cite{saunders2020progressive}. We provided 5 30-second videos of ground-truth skeletal sequences and corresponding photo-realistic videos to deaf participants, asking them to rate each video out of 5 for understandability. Synthesised videos were rated higher for comprehension, at 3.9 compared to 3.2 for skeletal sequences. This suggests that photo-realistic content is more understandable to a deaf signer than a skeleton sequence.

\subsection{Qualitative Evaluation} \label{sec:qual_experiments}

We show example generated photo-realistic frames in Fig. \ref{fig:Qual_Eval}, highlighting the production quality. We provide further qualitative evaluation in supplementary materials.

\section{Potential Negative Societal Impact} \label{sec:societal_impact}

We acknowledge the potential use of \ac{slp} technology to remove the reliance on human interpreters. However, we see this work as enabling a larger provision of signed content, especially where interpretation doesn't exist \cite{dickinson2017sign}. We also recognise the potential harm if this technology produced incorrect sign language content, particularly in emergency settings. Although this paper significantly advances the \ac{slp} field, we would like to state that \ac{slp} technology is still under development and should not yet be relied upon.

\section{Conclusion} \label{sec:conclusion}

Large-scale photo-realistic \ac{slp} is important to provide high quality signing content to deaf communities. In this paper, we proposed the first \ac{slp} method to achieve both large-scale signing and photo-realistic video generation. We proposed \FrameSelectionNet{}, which learns to co-articulate between dictionary signs by modelling the optimal temporal alignment to continuous sequences. Furthermore, we proposed \methodName{} to produce photo-realistic sign language videos. We proposed a novel keypoint-based loss function that improves the quality of hand synthesis, operating in the keypoint space to avoid issues caused by motion blur. 

We showed how our approach can scale to unconstrained domains of discourse and be understood by native signers, with considerable state-of-the-art \ac{ph14t} back translation performance. Additionally, we performed extensive user evaluation showing our approach increases the realism of interpolated dictionary signs, can be understood by native deaf signers and is overwhelmingly preferred to baseline methods. Finally, we showed that \methodName{} outperforms all baseline methods for quantitative metrics, human evaluation and native deaf signer comprehension.

Our approach is limited by the current performance of text to gloss translation for large-scale domains. Available gloss annotations are limited, making sign language translation tasks a low-resource machine translation task \cite{moryossef2021data}. Improvements on both architectures and datasets are required to compete with spoken language \ac{nmt} methods. For future work, we plan to tackle spatial co-articulation between dictionary signs.

\section*{Acknowledgments}

This project was supported by the EPSRC project ExTOL (EP/R03298X/1), SNSF project SMILE-II (CRSII5 193686) and EU project EASIER (ICT-57-2020-101016982). This work reflects only the authors view and the Commission is not responsible for any use that may be made of the information it contains. We thank Thomas Hanke and University of Hamburg for use of the \ac{mdgs} data. We thank SWISSTXT for the use of interpreter appearance data.

{\small
\bibliographystyle{ieee_fullname}
\bibliography{bibliography}
}

\newpage
\appendix
\appendixpage

\section*{\acf{mdgs} Translation Protocol}

In this appendix, we provide further details of our released translation protocols on the \acf{mdgs} dataset \cite{hanke2010dgs}. The public \ac{mdgs} linguistic corpus can be accessed at \url{https://www.sign-lang.uni-hamburg.de/meinedgs/}, containing 330 sequences of free-flowing discourse between two deaf participants, with each around 10 minutes in length.  Additionally, detailed spoken language transcripts, frame-level gloss annotations and 2D pose estimation sequences \cite{cao2018openpose} are provided. Discourse is centered around a wide variety of topics, age groups and format, with further details available on the mDGS website. 

To adapt the \ac{mdgs} corpus for use as a translation dataset, we segment the free-flowing discourse data into 40,230 segments of German sentences, sign gloss translations and respective sign language videos. Sequence segmentation was performed using spoken language sentence boundaries, with corresponding frame boundaries provided. The title of each segment (e.g. 1583882A-X) contains the title of the original discourse sequence as given in the \textit{Transcript} column (e.g. 1583882), the corresponding participant camera (A or B) and the position of the extracted segment in the original discourse sequence (a numerical value X).

\begin{table}[b!]
\centering
\resizebox{1\linewidth}{!}{%
\begin{tabular}{@{}p{1.6cm}|ccc|ccc@{}}
\toprule
\multicolumn{1}{c}{} & \multicolumn{3}{c}{Sign Gloss} & \multicolumn{3}{c}{German} \\
           & Train   & Dev    & Test   & Train  & Dev   & Test  \\  \midrule
segments   & 40,230 &  4,996 & 4,977 & \multicolumn{3}{c}{$\xleftarrow{\rule{2.0cm}{0pt}}same$} \\
frames     & 6,146,153 & 764,451 & 758,883 & \multicolumn{3}{c}{$\xleftarrow{\rule{2.0cm}{0pt}}same$} \\
vocab.     & 10,042 & 4,644 & 4,620 & 18,680  & 6,224  & 6,231  \\
tot. words & 215,392 & 26,855 & 26,969 & 389,427 & 48,376 & 48,551 \\
tot. OOVs  & - & 371 & 339 & - & 1,103 & 1,171 \\
singletons & 2,681 & - & -  & 8,909 & - & - \\
\bottomrule
\noalign{\smallskip} 
\end{tabular}}
\caption{Key statistics of the \acf{mdgsv} dataset split.}
\label{tab:mDGSv_Stats}
\end{table}

Table \ref{tab:mDGSv_Stats} and \ref{tab:mDGS_Stats} show detailed statistics of the \ac{mdgsv} and \ac{mdgs} protocols, respectively. Gloss variants used in \ac{mdgsv} give distinction between sign variants, with each containing the same meaning but with differing motion. We chose to retain these variants to provide more challenging baselines for the community. Further public annotation conventions are outlined in \cite{konrad2018public}, which we follow. Additionally, gloss frame alignments are provided as \textit{GLOSS/start-frame/stop-frame} (e.g. BUCHSTABE1/11/34). The translation protocols are publicly available at \url{https://github.com/BenSaunders27/meineDGS-Translation-Protocols}, detailing \textit{filename}, \textit{camera}, \textit{ger\_text}, \textit{gloss}, \textit{start\_time} and \textit{stop\_time}.

To use the \ac{mdgs} dataset for computational research, a licence must be obtained from the University of Hamburg\footnote{\url{https://www.sign-lang.uni-hamburg.de/meinedgs/}}. Release of these protocols does not imply permission for use or provide a license. Written permission is required from the dataset owner. Please adhere to the data ownership policies and ensure you have the correct rights of use.

\begin{table}[]
\centering
\resizebox{1\linewidth}{!}{%
\begin{tabular}{@{}p{1.6cm}|ccc|ccc@{}}
\toprule
\multicolumn{1}{c}{} & \multicolumn{3}{c}{Sign Gloss} & \multicolumn{3}{c}{German} \\
           & Train   & Dev    & Test   & Train  & Dev   & Test  \\  \midrule
segments   & 40,230 &  4,996 & 4,977 & \multicolumn{3}{c}{$\xleftarrow{\rule{2.0cm}{0pt}}same$} \\
frames     & 6,146,153 & 764,451 & 758,883 & \multicolumn{3}{c}{$\xleftarrow{\rule{2.0cm}{0pt}}same$} \\
vocab.     & 4,337 & 2,490 & 2,487 & 18,680  & 6,224  & 6,231  \\
tot. words & 215,392 & 26,855 & 26,969 & 389,427 & 48,376 & 48,551 \\
tot. OOVs  & - & 118 & 112 & - & 1,103 & 1,171 \\
singletons & 778 & - & -  & 8,909 & - & - \\
\bottomrule
\noalign{\smallskip} 
\end{tabular}}
\caption{Key statistics of the \acf{mdgs} dataset split.}
\label{tab:mDGS_Stats}
\end{table}

\end{document}